\documentclass[a4paper,conference]{IEEEtran}

\usepackage{color}
\usepackage{graphicx}
\usepackage{amssymb}
\usepackage{makecell}
\usepackage{caption}
\usepackage{subcaption}
\usepackage{comment}
\usepackage{multirow}
\usepackage{hyperref}

\newcommand{\gray}{\emph{Gray} }
\newcommand{\grayd}{\emph{GrayD} }
\newcommand{\rgb}{\emph{RGB} }
\newcommand{\rgbd}{\emph{RGBD} }

\ifCLASSINFOpdf
\else
\fi
\begin{document}
%
\title{DepthFake: a depth-based strategy for detecting Deepfake videos}

\author{Luca Maiano, Lorenzo Papa, Ketbjano Vocaj and Irene Amerini  \\ Sapienza University of Rome, Italy\\
Email: [maiano, papa, amerini]@diag.uniroma1.it}


%


\maketitle

\begin{abstract}
Fake content has grown at an incredible rate over the past few years. The spread of social media and online platforms makes their dissemination on a large scale increasingly accessible by malicious actors. In parallel, due to the growing diffusion of fake image generation methods, many Deep Learning-based detection techniques have been proposed. Most of those methods rely on extracting salient features from RGB images to detect through a binary classifier if the image is fake or real.

In this paper, we proposed DepthFake, a study on how to improve classical RGB-based approaches with depth-maps. The depth information is extracted from RGB images with recent monocular depth estimation techniques. Here, we demonstrate the effective contribution of depth-maps to the deepfake detection task on robust pre-trained architectures. The proposed \rgbd approach is in fact able to achieve an average improvement of $3.20\%$ and up to $11.7\%$ for some deepfake attacks with respect to standard RGB architectures over the FaceForensic++ dataset. 
\end{abstract}


%
\IEEEpeerreviewmaketitle

\section{Introduction}

Recent advances in artificial intelligence enable the generation of fake images and videos with an incredible level of realism. The latest deepfakes on the war in Ukraine\footnote{\href{https://www.bbc.com/news/technology-60780142}{https://www.bbc.com/news/technology-60780142}} have clarified the need to develop solutions to detect generated videos. This is just the latest example in a long series of realistic deepfake videos that have hit public figures in the last few years. Deepfakes have become a real threat, and the techniques for generating this content are advancing at an incredible speed. The term deepfake originally came from a Reddit user called “deepfakes” that in 2017 used off-the-shelf AI tools to paste celebrities' faces onto pornographic video clips. Today, the term typically has a broader meaning and refers to every kind of face manipulation technique based on artificial intelligence and computer graphics. 

Automatic face manipulations can be divided in two main categories: \emph{face swap} or \emph{reenactment}~\cite{10.1145/3306346.3323035,Thies_2016_CVPR}. Face swap applies some source characteristics onto a target face. Reenactment is a more sophisticated technique that enables the reenactment of a target person while maintaining its natural aspect. Dealing with such a heterogeneous range of manipulations is the main challenge for current deepfake video detectors. In fact, video content creation techniques are still improving month by month, which makes video content detection an even more complex problem, as for each new detector there is always a new generation method that is more realistic than the previous one.

\begin{figure}[t]
    \centering
    \begin{subfigure}{0.24\linewidth}
        \centering
        \includegraphics[width=\linewidth]{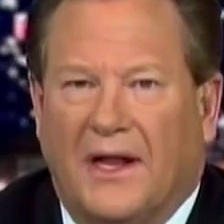}
        \hfill
        \includegraphics[width=\linewidth]{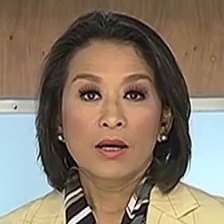}
        \hfill
        \includegraphics[width=\linewidth]{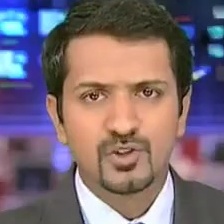}
        \caption{Original face}
        \label{fig:rgb1}
    \end{subfigure}
    \begin{subfigure}{0.24\linewidth}
        \centering
        \includegraphics[width=\linewidth]{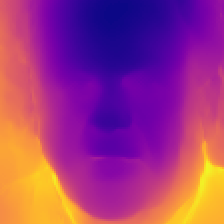}
        \hfill
        \includegraphics[width=\linewidth]{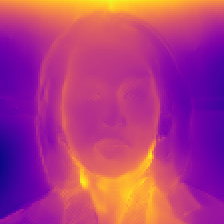}
        \hfill
        \includegraphics[width=\linewidth]{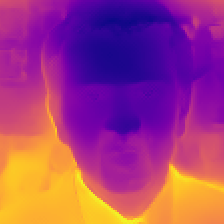}
        \caption{Real depth}
        \label{fig:rgb1_depth}
    \end{subfigure}
    \begin{subfigure}{0.24\linewidth}
        \centering
        \includegraphics[width=\linewidth]{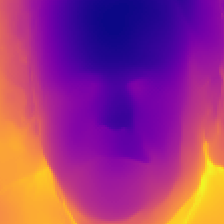}
        \hfill
        \includegraphics[width=\linewidth]{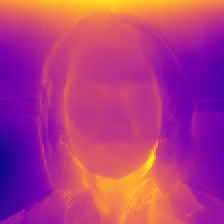}
        \hfill
        \includegraphics[width=\linewidth]{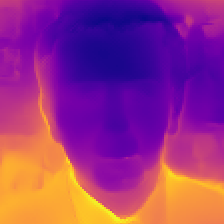}
        \caption{Fake depth}
        \label{fig:rgb1_fake_depth}
    \end{subfigure}
    \begin{subfigure}{0.24\linewidth}
        \centering
        \includegraphics[width=\linewidth]{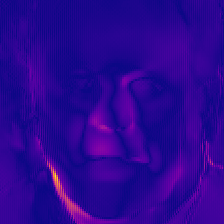}
        \hfill
        \includegraphics[width=\linewidth]{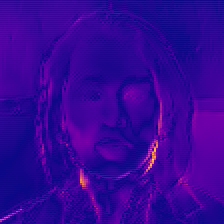}
        \hfill
        \includegraphics[width=\linewidth]{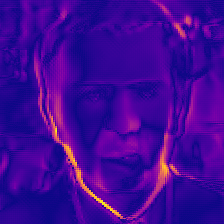}
        \caption{Difference}
        \label{fig:rgb1_difference}
    \end{subfigure}
    \caption{Some example inconsistencies introduced in the depth map of manipulated faces. Deepfake faces tend to have less details than the original ones.}
    \label{fig:examples}
\end{figure}

In this paper, we analyze the depth inconsistencies introduced by face manipulation methods. Unlike methods that analyze either imaging pipelines (e.g., PRNU noise~\cite{lukas2005determining}, specifications~\cite{huh2018fighting}), encoding approaches (e.g., JPEG compression patterns~\cite{barni2017aligned}), or image fingerprint methods~\cite{yu2019attributing}, our study analyzes the alteration introduced by the manipulation on RGB and depth features. These spatial features contain semantic information that has the advantage of being more easily interpretable and robust to strong compression operations. With these strengths, \emph{semantic features} can help solve two major challenges with deepfake detection. On the one hand, the lack of explainable detectors, which do not limit themselves to classifying the contents as true or false but allow us to understand what information led to a certain decision. On the other hand, these detectors should be robust to detect fake videos even when some low-level information gets destroyed by compression algorithms. This is particularly important when a video is disseminated on social networks and published several times by different users. In fact, most platforms often reduce the quality and resolution of the video, which can weaken many deepfake detectors. To analyze these semantic features, in this paper we propose to extract the depth of the face with a monocular-based estimation method that is concatenated to the RGB image. We than train a network to classify each video frame as \emph{real} or \emph{fake}. These two modalities enable the analysis of semantic inconsistencies in each frame by investigating color and spatial irregularities. 

To demonstrate the effectiveness of our \emph{DepthFake} method, we conduct extensive experiments on different classes included in the FaceForensics++~\cite{Rossler2019ICCV} dataset. In our experiments, we demonstrate the effectiveness of our method by introducing a vanilla \rgb baseline and demonstrating that adding depth information allows us to systematically improve detection performance. In summary, the main contributions of this paper are threefold as below.
\begin{itemize}
    \item We analyze the importance of depth features and show that they consistently improve the detection rate. \emph{To the best of our knowledge}, this is the first work that analyses the depth inconsistencies left by the deepfake creation process. Figure~\ref{fig:examples} shows some example of depth inconsistencies.
    \item We investigate the contribution of the RGB data and show that a simple RGB-to-grayscale conversion can still lead to acceptable or even higher results in some experiments. We hypothesize that there are semantic features in this conversion that still allow good detection despite the reduction of input channels.
    \item We conduct preliminary experiments on inference times required by one of the most used convolutional neural networks on several hardware configurations. The increasingly massive adoption of streaming and video conferencing applications brings the need to develop deepfake detection solutions in \emph{real-time}. With this work, we propose some experiments to analyze the impact of using multiple channels such as depth or grayscale features on inference times. Our aim is to analyze the impact of our multi-channel model on inference time. These first experiments are a valuable baseline for future developments and studies.
\end{itemize}

The remainder of this paper is organized as follows: Section~\ref{sec:related-works} presents state-of-the-art methods for deepfake generation, detection, and  single image depth estimation. The proposed detection technique is explained in Section~\ref{sec:method} as regards the feature extraction and the classification phase. Section~\ref{sec:results} reports the experimental results. Finally, Section~\ref{sec:conclusions} concludes the paper with insights for future works.

\section{Related Works}
\label{sec:related-works}

In this section, we examine the state-of-the-art by comparing the contribution of our method to the others. The section begins by reviewing the deepfake detection techniques and ends with an overview of the state-of-the-art methods for monocular depth estimation.

\subsection{Deepfake Detection}
\label{subsec:deepfake_detection}
There are four common strategies to create a deepfake video: (i) face identity swap, (ii) facial attribute or (iii) expression modification and (iv) entire image synthesis. These methods are used in combination with 3D models, AutoEncoders, or Generative Adversarial Networks to forge the fake image that is then blended into the original one. Recent studies~\cite{9675329,yu2018attributing} demonstrate the possibility of detecting the specific cues introduced by this fake generation process. Some methods train deep neural networks to learn these features. Some other methods analyze the semantics of the video to search for inconsistencies. More recently, some researchers have come up with the idea of testing whether a person's behavior in a video is consistent with a given set of sample videos of this person, which we call re-identification for the sake of brevity. Our work falls between the first two approaches and examines RGB and depth features. In the remainder of this section, we discuss the most related detection approaches.\bigbreak

\textbf{Learned featues.}  Most efforts focus on identifying features that can be learned from a model. The study from Afchar et al.~\cite{afchar2018mesonet} was one of the first approaches for deefake detection based on supervised learning. This method focuses on mesoscopic features to analyze the video frames by using a network with a low number of layers. Rossler et al.~\cite{Rossler2019ICCV} investigate the performance of several CNN architectures for deepfake video detection and show that deeper networks are more effective for this task. Also, the attention mechanism  introduced by Dang et al.~\cite{Dang_2020_CVPR} improves the localization of the manipulated regions. Following this research trend, our method adds a step that can be used in conjunction with other techniques to improve the detection of fake videos.

In addition to spatial inconsistencies, fake artifacts arise along the temporal direction and can be used for forensic investigation. Caldelli et al.~\cite{CALDELLI202131} exploit temporal discrepancies through the analysis of optical flow features. Guera et al.~\cite{guera2018deepfake}  propose the use of convolutional Long Short Term Memory (LSTM) network while Masi et al.~\cite{978-3-030-58571-6_39} extract RGB and residual image features and fuse them into an LSTM. Like for the spatial domain, the attention also improves the detection of temporal artifacts~\cite{zi2020wilddeepfake}. In this preliminary work, we do not consider the temporal inconsistencies of the depth, although we believe that this analysis can further improve the performance of a model based on these features and we leave this investigation for future works.

Moreover, data augmentation strategies constitute another ingredient to improve the detection and generalization~\cite{9675329}. Top performers in the recent DeepFake's detection challenge~\cite{dolhansky2020deepfake} use extensive augmentation techniques. In particular, the augmentations based on the cut-off on some specific parts of the face have proved to be particularly effective~\cite{du2020towards,bonettini2021video}.\bigbreak

\textbf{Semantic featues.} Several studies look at specific semantic artifacts of the fake video. Many works have focused on biological signals such as blinking~\cite{liy2018exposingaicreated} (which occurs with a certain frequency and duration in real videos), heartbeat~\cite{qi2020deeprhythm} and other biological signals to find anomalies in the spatial and temporal direction. Other approaches use inconsistencies on head pose~\cite{yang2019exposing} or face warping~\cite{li2018exposing} artifacts to detect fake content. 

Recently, multiple modalities have also been proved very effective for deepfake detection. Mittal et al.~\cite{mittal2020emotions} use audio-visual features to detect emotion inconsistencies in the subject. Zhou et al.~\cite{zhou2021joint} use a similar approach to analyze the intrinsic synchronization between the video and audio modalities. Zhao et al.~\cite{zhao2021multi} introduce a multimodal attention method that fuses visual and textual features. 
While our method does not fall directly into this category, we still incorporate some semantic information into our model. In fact, we use CNN architectures pre-trained on ImageNet~\cite{deng2009imagenet}, which still leads our model to extract high-level features on image semantics. Furthermore, anomalies in depth can be considered semantic features, as a real face should have features in three dimensions. If there are anomalies in this sense, that is a "flat" face, this would introduce a semantic inconsistency.\bigbreak

\textbf{Re-identification features.} Re-identification methods distinguish each individual by extracting some specific biometric traits that can be hardly reproduced by a generator~\cite{agarwal2019protecting, agarwal2020detecting, agarwal2020detecting}. The first work of this kind was introduced by Agarwal et al.~\cite{agarwal2019protecting} and exploits the distinct patterns of facial and head movements of an individual to detect fake videos. In another work~\cite{agarwal2020detecting}, the same research group studied the inconsistencies between the mouth shape dynamics and a spoken phoneme. Cozzolino et al.~\cite{cozzolino2021id} introduced a method that extracts facial features based on a 3D morphable model and focuses on temporal behavior through an adversarial learning strategy. Another work~\cite{arxiv.2204.03083} introduces a contrastive method based on audio-visual features for person-of-interest deepfake detection.\bigbreak

\begin{figure*}[t]
    \centering
    \includegraphics[width=0.9\linewidth]{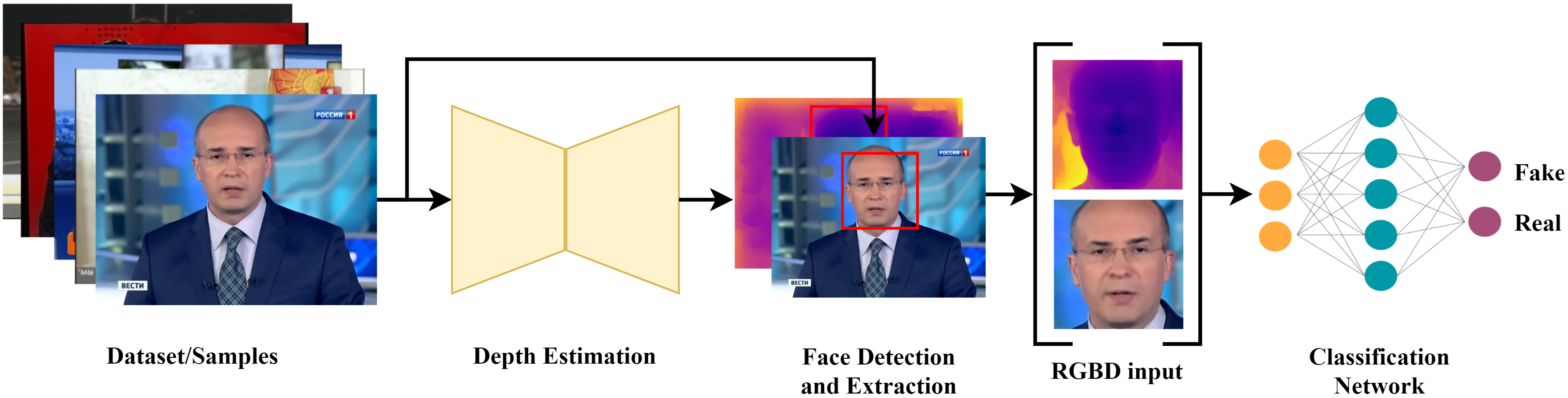}
    \caption{Pipeline of the proposed method. In the fist step, we estimate the depth for each frame. Then, we extract the face and crop the frame and depth map around the face. In the last step, we train a classifier on \rgbd input features.}
    \label{fig:pipeline}
\end{figure*}

\subsection{Monocular Depth Estimation}
\label{subsec:monocular_depth_estimation}
Monocular depth estimation (MDE) is a fundamental task in computer vision where a per-pixel distance map is reconstructed from a single RGB input image. Unlike other passive depth perception methodologies such as the binocular stereo vision and the multi-view systems, this depth perception approach does not impose any geometric constraints due to the use of a single camera, thus leading to a multitude of possible applications such as scene understanding and robotics.

Multiple works have been proposed to tackle the problem. Recent works are mainly divided into two trends: the transformers and deep convolutional architectures that aim to predict a depth-map with the highest estimation accuracy~\cite{adabins, DPT, BinsFormer}, while the other trend focuses their attention on lightweight architectures~\cite{speed, depthlite_0, depthlite_1} to infer at high frame rates on embedded and mobile devices.
Moreover, depth information can be successfully integrated with RGB data, i.e. into an \rgbd approach, to obtain notable improvements in other challenging tasks. He et al.~\cite{He2021SOSDNetJS} propose SOSD-Net, an multi-task learning architecture for semantic object segmentation and monocular depth estimation enforcing geometric constraint while sharing common features between the two tasks. Vége et al.~\cite{posedepth} propose a two step architecture based on 2D pose estimation and monocular depth images for the 3D pose estimation task. 

Similarly to the just introduced works, in this study we take advantage of depth-maps information to tackle the deepfake detection task. Moreover, due to the lack of depth data we use the pre-trained monocular depth estimation model proposed by Khan et at.~\cite{face_depth_ket} to extract depth face-images from deepfake datasets.  Khan et at. propose a convolutional encoder-decoder network trained on estimating monocular depth of subjects which appear closer to the camera, i.e. with a limited depth range that goes from 0.3 to 5.0 meters; with this setup we are able to explore and understand the depth channel contribution for the deepfake detection task.

\section{Proposed Method}
\label{sec:method}

In this section, we introduce \emph{DepthFake}. Our system is structured in two steps. First, we estimate the depth of the entire image through the FaceDepth model proposed by Khan et al.~\cite{face_depth_ket}. This model is pre-trained to estimate the depth of human faces. Next, in the second phase we extract a $224\times224$ patch of the subject's face from both the RGB image and depth map. This step allows extracting the face without having to resize the image, as resizing may eliminate precious details useful for classification. Finally, we train a convolutional neural network to classify whether the content is fake or real.

In Sections~\ref{subsec:depth} and~\ref{subsec:depthfake}, we delve into the two modules with further details on the system represented in Figure~\ref{fig:pipeline}, while in Section~\ref{subsec:implementation} we discuss about the implementation details.

\subsection{Depth Estimation}
\label{subsec:depth}
Depth estimation is at the heart of our method. In fact, we hypothesize that the process of generating deepfakes introduces depth distortions on the subject's face. Therefore, the first step in the proposed pipeline is extracting the depth of the face. As explained in Section~\ref{sec:related-works}, we can estimate the depth of an image through a monocular depth estimation technique. However, since there are no deepfake datasets containing ground-truth depth information, we propose to use a pre-trained model. 

Current deepfakes are usually created with a foreground subject. Therefore, we adopt the FaceDepth~\cite{face_depth_ket}, a network trained to estimate the distance of the subjects captured by the camera. The model is trained on synthetic and realistic 3D data where the camera is set at a distance of thirty centimeters from the subject and the maximum distance of the background is five meters. This allows us to discriminate facial features by obtaining fine-grained information on the depth of each point of the face. The model has an encoder-decoder structure and consists of a MobileNet~\cite{arxiv.1704.04861} network that works as a feature extractor, followed by a series of upsampling layers and a single pointwise layer. The network was trained to receive $480\times640$ input images and output a $240\times320$ depth map. The estimated map constitutes one of the four input channels of our deepfake discriminator.

\subsection{Deepfake Detection}
\label{subsec:depthfake}
The second module of our system concatenates the estimated depth map to the original RGB image. Since the alterations introduced by deepfakes are usually more significant on the subject's face, we crop $224 \times 224$ pixels to extract the face from the rest of the image. The result of this concatenation generates an \rgbd tensor $x \in \mathbb{R}^{224 \times 224 \times 4}$, which constitutes the input of our classification network.

In the last step of our method, we train a neural network to classify real and fake video frames. In terms of architecture, we use an Xception~\cite{Chollet_2017_CVPR} network pre-trained on ImageNet~\cite{deng2009imagenet}. Since we are using a network pre-trained on classical RGB images, i.e. the one used for ImageNet, the addition of the depth channel as forth input creates the need to adapt and modify the initial structure of the original network to handle this type of data while guaranteeing the correct weights initialization. Therefore, if we randomly initialized an input layer with 4 channels, this would end up heavily affecting the weights learned during the pre-training phase. To solve this problem, we decided to add an additional channel obtained by calculating the average of the three original input channels from the pre-trained model. This change makes the training more stable and allows the model to converge towards the optimum fastly. Consequently, we have chosen to use this initialization method for all the experiments. In addition to this, there is a further problem to be taken into consideration. The values contained in the depth channel range from 0 to 5000, which is the range in which the depth estimation module has been pre-trained. Linking this channel to the RGB channel without normalizing these values would end up causing numerical instability and would heavily cancel the RGB contribution. To handle this problem we normalize the depth channel values in the range $0$--$255$.

In terms of augmentations, we apply flipping and rotation. While it has been shown that the strongest boost based on compression and blurring generally improves performance for this task, we decide to keep the augment strategy as simple as possible to avoid altering the information provided by the depth channel.

Added to this, we investigate the contribution of the RGB color model for deepfake detection when paired to the depth information. To this end, we train our system on 2-channel grayscale plus depth input data ($x \in \mathbb{R}^{224 \times 224 \times 2}$). The results reported in Section~\ref{sec:results}, show that a system trained on depth and grayscale features achieves acceptable or higher results than \rgbd input data and superior results compared to standard RGB inputs. In this configuration, the network may assign a greater contribution to the depth channel, thus reducing the importance of the information contained in the RGB space. While this is not the goal of this work, it allows us to analyze the impact of a different number of input channels on model inference times, which can be extremely important for real-time applications.

\begin{table*}[ht!]
    \centering
    \caption{
        Accuracy on Deepfake (DF), Face2Face (F2F), FaceSwap (FS), NeuralTexture (NT) and the full dataset (FULL) with RGB and RGBD inputs. Bold represents the best configuration of each backbone and underlined accuracies represent the best value over each class. In brackets we indicate the percentage difference added by the depth.
    }
    \label{tab:binary_classification}
    \begin{tabular}{ l | c | c | c | c | c | c }

        \multicolumn{1}{c |}{}  & \multicolumn{2}{c |}{ResNet50} & \multicolumn{2}{c|}{MobileNet--V1} & \multicolumn{2}{c}{XceptioNet}\\
        \cline{2-7}
        \multicolumn{1}{c|}{} & RGB & RGBD & RGB & RGBD & RGB & RGBD \\
        \hline
        DF & $93.91\%$ & \makecell{$\mathbf{94.71}\%$ \\ $(+0.8)$}  & $95.14\%$ & \makecell{$\mathbf{95.86}\%$ \\ $(+0.72)$} & $97.65\%$ & \makecell{$\underline{\mathbf{97.76}}\%$ \\ $(+0.11)$}\\
        \hline
        F2F & $96.42\%$ & \makecell{$\mathbf{96.58}\%$ \\ $(+0.16)$} & $97.07\%$ & \makecell{$\underline{\mathbf{98.44}}\%$ \\ $(+1.37)$} & $95.82\%$ & \makecell{$\mathbf{97.41}\%$ \\ $(+1.59)$}\\ 
        \hline
        FS & $97.14\%$ & \makecell{$\mathbf{96.95}\%$ \\ $(-0.19)$} & $97.12\%$ & \makecell{$\mathbf{97.87\%}$\\ $(+0.75)$} & $97.84\%$ & \makecell{$\underline{\mathbf{98.80}}\%$ \\ $(+0.96)$}\\
        \hline
        NT & $75.81\%$ & \makecell{$\mathbf{77.42}\%$ \\ $(+1.61)$} & $70.47\%$ & \makecell{$\mathbf{82.26}\%$ \\ $(+11.79)$} & $76.87\%$ & \makecell{$\underline{\mathbf{85.09}}\%$ \\ $(+8.22)$}\\
        \hline
        FULL & $85.68\%$ & \makecell{$\mathbf{90.48\%}$ \\ $(+4.80)$} & $90.60\%$ & \makecell{$\mathbf{91.12}\%$ \\ $(+0.52)$} & $86.80\%$ & \makecell{$\underline{\mathbf{91.93}}\%$ \\ $(+5.13)$}\\
       
    \end{tabular}
    \label{tab:accforg}
\end{table*}

\begin{table}[ht!]
    \centering
    \caption{
        Accuracy of the Xception-based model on video level for Deepfake (DF), Face2Face (F2F), FaceSwap (FS), NeuralTexture (NT) and the full dataset (FULL) with RGB and RGBD inputs. 
        In brackets we indicate the percentage difference added by the depth, while bold represents the best configuration.
    }
    \label{tab:video_level}
    \begin{tabular}{ l | c | c }
        \multicolumn{1}{c|}{} & RGB & RGBD \\
        \hline
        DF & $97.25\%$ & \makecell{$\mathbf{98.85}\%$ \\ $+(1.60)$} \\
        \hline
        F2F & $97.50\%$ & \makecell{$\mathbf{98.75}\%$ \\ $+(2.25)$}\\ 
        \hline
        FS & $98.00\%$ & \makecell{$\mathbf{98.75}\%$ \\ $+(0.75)$}\\
        \hline
        NT & $81.25\%$ & \makecell{$\mathbf{88.00}\%$ \\ $+(6.25)$} \\
        \hline
        FULL & $87.00\%$ & \makecell{$\mathbf{93.00}\%$ \\ $+(6.00)$} \\
    \end{tabular}
    \label{tab:accvideo}
\end{table}

\subsection{Implementation Details}
\label{subsec:implementation}
We implement the proposed study using the \emph{TensorFlow}\footnote{https://www.tensorflow.org/} API and train the system on an NVIDIA GTX 1080 with 8GB of memory. In all the experiments we use ADAMAX as optimizer
with the following setup: a starting learning rate equal to $0.01$ with $0.1$ decay, $\beta_1=0.9$, $\beta_2=0.999$, and $\epsilon=1e^{-07}$. The training process is performed for $25$ epochs with a batch size of $32$. We set the input resolution of the architectures equal to $224\times224$ while cropping the original input image around the face. The face detection and extraction is performed with \emph{dlib}\footnote{http://dlib.net/}. The loss function chosen for the training process is the Binary Crossentropy, a widely employed function for classification tasks; its mathematical formulation is reported in Equation~\ref{eq:loss}, where we indicate with $\hat{y}_i$ the predicted sample and with $y_i$ the target one.
\begin{equation}
    Loss = - \frac{1}{2} \sum_{i=1}^{2} y_i \cdot log \hat{y}_i + (1-y_i) \cdot log(1 - \hat{y}_i)
    \label{eq:loss}
\end{equation}

Once the training phase has been completed, we compare the inference times between the different models and input channels; we report those values in milliseconds (ms) in Section~\ref{sec:results}. 

\section{Results}
\label{sec:results}
In this section we report the experiments and evaluations that have been conducted. We propose a comparison with some well established CNN networks and show the first results on the inference times of the model that will be deepened in future studies. We evaluate our model on the FaceForensic++~\cite{Rossler2019ICCV} dataset, which allows us to evaluate the importance of depth features on the most common strategies to create a deepfake. This dataset is composed of 1000 original video sequences that have been manipulated with four face forgeries, namely: Deepfake (DF), Face2Face (F2F), FaceSwap (FS) and NeuralTexture (NT). For all experiments, we split the dataset into a fixed training, validation, and test set, consisting of 90\%, 5\%, and 5\% of the videos, respectively. 

\subsection{Deepfake detection}

We begin our experiments by analyzing the effectiveness of the proposed solution in identifying deepfakes. We train our system to solve a binary classification task on individual video frames. We evaluated multiple variants of our approach by using different state-of-the-art classification methods. In addition, we show that the classification based on the Xception~\cite{Chollet_2017_CVPR} architecture outperforms all other variants in detecting fakes, as previously demonstrated in other works~\cite{Rossler2019ICCV,gragnaniello2021gan}.

\begin{table*}[ht!]
    \centering
    \caption{
        Floating point operations (FLOPS) and average frame per second (FPS) inference frequency over different platforms. For the GrayD and RGBD we indicate in brackets the difference with respect to the Gray and RGB models respectively.
    }
    \label{tab:binary_classification}
    \begin{tabular}{ l | c | c | c | c | c }
        \multicolumn{1}{c |}{}  & ARM Cortex CPU & Nvidia GTX 1080 & Nvidia Titan V  & Nvidia RTX 3090 & Giga \\
        \multicolumn{1}{c|}{} & [fps] & [fps]  & [fps]  & [fps] & FLOPS \\
        \hline
        Gray & 0.497 & 28.33  & 25.02  & 20.41 & 9.19\\ 
        \hline
        GrayD &  \makecell{0.496 \\ (-0.001)} & \makecell{27.91 \\ (-0.42)} & \makecell{25.66 \\ (-0.26)} & \makecell{19.96 \\ (-0.45)}  & \makecell{9.20 \\ (+0.01)}\\ 
        \hline
        RGB & 0.499 & 27.89  & 26.54  & 21.27 & 9.211\\ 
        \hline
        RGBD & \makecell{0.495 \\ (-0.005)} & \makecell{27.94 \\ (-0.05)}  & \makecell{23.77 \\ (-2.77)}  & \makecell{19.74 \\ (-1.53)} & \makecell{9.218 \\ (+0.007)}\\ 
        \hline
    \end{tabular}
    \label{tab:fps}
\end{table*}

First, we evaluate the effectiveness of the main backbones that are popular for deepfake detection: ResNet50~\cite{resnet}, MobileNet~\cite{arxiv.1704.04861}, and XceptionNet~\cite{Chollet_2017_CVPR}. 
Table~\ref{tab:binary_classification} compares the results of our experiments with all our configurations. As shown in other studies~\cite{Rossler2019ICCV, 9675329}, the Xception network achieves the best performance on all backbones. The results show that the depth input channel always improves the model's performance in all configurations. Added to this, it is interesting to note that the MobileNet is slightly inferior to the Xception and outperforms the deeper ResNet50. This is a notable result when considering the goal of reducing inference times for real-time applications. While this is not the main contribution of this work, we still consider it an encouraging result for future developments. 

\begin{figure}[t]
    \centering
    \includegraphics[width=0.7\linewidth]{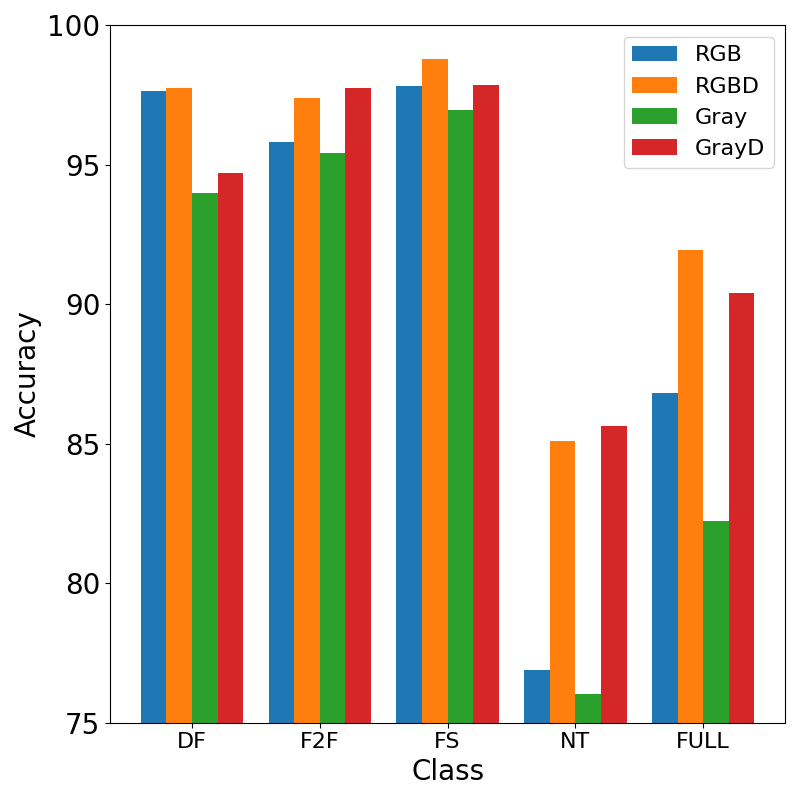}
    \caption{Accuracy on Deepfake (DF), Face2Face (F2F), FaceSwap (FS), NeuralTexture (NT) and all classes in the dataset (FULL) with RGB, RGBD, Gray and GrayD inputs.}
    \label{fig:tests}
\end{figure}

Next, to have a complete overview of the depth contribution, we compare the Xception's performances through the following four setups.
\begin{itemize}
    \item \textbf{RGB}. The baseline on which the different backbones have been trained using only the RGB channels.
    \item \textbf{Gray}. The backbone trained on grayscale image solely.
    \item \textbf{RGBD}. The model that is trained on 4-channel inputs based on the composition of the RGB and depth channels.
    \item \textbf{GrayD}. The configuration that is trained on 2-channel inputs composed of grayscale and depth channels.
\end{itemize}
As shown in Figure~\ref{fig:tests}, the results reveal a consistent advantage of the \rgbd and \grayd configurations over other \rgb and \gray ones. In particular, this advantage is more evident in the NeuralTexture class, which is also the most difficult class to recognize among those analyzed. 
For the \grayd configuration, the results are comparable or in some cases even higher than the performance of the model trained on \rgbd data. These results confirm our initial hypothesis that depth can make a significant contribution to the detection of deepfakes. In the \rgbd configuration, the model learns to reduce the contribution of the information contained in the RGB channels, while in the \grayd configuration, a lot of irrelevant information has already been removed, allowing the model to obtain good results with fewer input channels. This result suggests that depth in this case adds a more relevant contribution to classification than color artifacts. Similar observations can be made by analyzing the results at the video level shown in Table~\ref{tab:video_level}. In this case, the performances were measured by predicting the most voted class for each video.


\subsection{Preliminary studies on inference time}

We conclude our study by presenting preliminary results on the inference times of the solution we have introduced. To the best of our knowledge, we are the first to analyze this aspect in detecting deepfakes. The inference time is of fundamental importance when considering the scenarios in which it is useful to detect a fake video in real-time. 
To do this we analyze the impact of using a different number of input channels on our system. Our aim, for this work,
is to analyze the inference times of our model to understand if the different configurations we have introduced have an impact on this aspect or not. Specifically, in Table~\ref{tab:fps} we report the floating point operations (FLOPS) and average frame per second (fps) of the Xception-based model on four different hardware platforms. The results suggest that the higher number of channels has a minimal impact over the inference time with an average $0.68$ reduction of frame per second.
The depth estimation step is not included in these computations; instead, only the facial extraction and deepfake detection stages are measured. As mentioned, at this stage we are only interested in studying the differences introduced by different number of input channels. Additionally, it is worth noting that even if we do not consider depth estimation in our measurements, there are numerous approaches for real-time monocular depth estimation that might be used for this phase~\cite{speed, depthlite_0, depthlite_1}. 

Based on these results, we can draw some consideration that can trace the right path to design a deepfake detector in real time. The first is that models like the Xception tend to be more effective at detecting fakes. This could suggest that the use of a lightweight network with layers inspired by this architecture could allow to obtain lower inference times while maintaining satisfactory performance. The second is that integrating features such as depth can improve the detection of fakes without affecting too much on the frames per second that the model can process. This aspect will be deepened in subsequent works.

\section{Conclusions and future work}
\label{sec:conclusions}
In this paper, we assessed the importance of depth features for deepfake detection. From our experiments, we can conclude that current deepfake detection methods can be supported through depth estimation features, which added to RGB channels can help increase the performance of the detectors. The results of this study will be investigated in the future with the aim of analyzing the generalization capacity of these features with respect to deepfake generation techniques that have not been seen in training and also evaluating the possibility of analyzing the variations in depth estimation over time. Moreover, we plan to test our model on other datasets than FaceForensics++.

In addition to that, in this work we have proposed the first studies on the impact of using different types of inputs with respect to inference times. These preliminary results will be a valuable reference basis for future work, which will investigate the limitations of current methods for real-time deepfake detection to design lighter strategies. The first results, suggest that even with shallower networks such as MobileNet it is still possible to obtain good performance. By mixing some design strategies like Xception-based layers and depth features, could be possible to designer stronger and lighter real-time methods.

\section{Acknowledgments}
This work has been partially supported by the Sapienza University of Rome project RM12117A56C08D64 2022-2024.

\bibliographystyle{IEEEtran}
\bibliography{main}

%



\end{document}